\documentclass[runningheads]{llncs}
\usepackage[T1]{fontenc}
\usepackage{amsmath} 
\usepackage{amssymb} 
\usepackage{booktabs}  
\usepackage{multirow}  
\usepackage{graphicx}  
\usepackage{float}
\usepackage[hidelinks]{hyperref}
\usepackage{marvosym}

\begin{document}
\title{Visually-Guided Controllable Medical Image Generation via Fine-Grained Semantic Disentanglement}
\titlerunning{Visually-Guided Generation via Semantic Disentanglement} 

\author{Xin Huang\inst{1,2}\textsuperscript{,$\dagger$} \and
Junjie Liang\inst{1,2}\textsuperscript{,$\dagger$} \and
Qingshan Hou\inst{1,2} \and
Peng Cao\inst{1,2,3}\textsuperscript{(\Letter)} \and
Jinzhu Yang\inst{1,2,3} \and
Xiaoli Liu\inst{4} \and
Osmar R. Zaiane\inst{5}}

\authorrunning{X. Huang et al.}

\institute{Computer Science and Engineering, Northeastern University, Shenyang, China \and
Key Laboratory of Intelligent Computing in Medical Image of Ministry of Education, Northeastern University, Shenyang, China \and
National Frontiers Science Center for Industrial Intelligence and Systems Optimization, Shenyang, China\\
\email{caopeng@cse.neu.edu.cn} \and
AiShiWeiLai AI Research, China \and
Amii, University of Alberta, Edmonton, Alberta, Canada}
    
\maketitle
{\let\thefootnote\relax\footnotetext{\textsuperscript{$\dagger$} The two authors contribute equally to this work.}}
%
\begin{abstract}
Medical image synthesis is crucial for alleviating data scarcity and privacy constraints. However, fine-tuning general text-to-image (T2I) models remains challenging, mainly due to the significant modality gap between complex visual details and abstract clinical text. In addition, semantic entanglement persists, where coarse-grained text embeddings blur the boundary between anatomical structures and imaging styles, thus weakening controllability during generation. To address this, we propose a Visually-Guided Text Disentanglement framework. We introduce a cross-modal latent alignment mechanism that leverages visual priors to explicitly disentangle unstructured text into independent semantic representations. Subsequently, a Hybrid Feature Fusion Module (HFFM) injects these features into a Diffusion Transformer (DiT) through separated channels, enabling fine-grained structural control. Experimental results in three datasets demonstrate that our method outperforms existing approaches in terms of generation quality and significantly improves performance on downstream classification tasks. The source code is available at \url{https://github.com/hx111/VG-MedGen}.

\end{abstract}
%
%
\section{Introduction}
High-quality annotated medical imaging datasets form the foundation for developing robust artificial intelligence (AI)-assisted diagnostic systems. However, acquiring such data remains challenging due to the scarcity of rare medical cases, the high cost of expert annotations, and strict privacy regulations\cite{shobayo2025developments,koutsoubis2025privacy,guo2025maisi}. To mitigate these challenges, Generative AI, particularly Latent Diffusion Models (LDM) \cite{rombach2022high} and their Transformer-based variants, such as DiT \cite{peebles2023scalable}, have gradually emerged as an effective approach to data enhancement. By conditioning themselves on textual descriptions, these models are able to synthesize diverse medical images, offering new opportunities to alleviate long-tailed medical distributions \cite{mao2025medsegfactory,huang2025vap,yuan2024adapting}.

\begin{figure}[t]
    \centering
    \includegraphics[width=\textwidth]{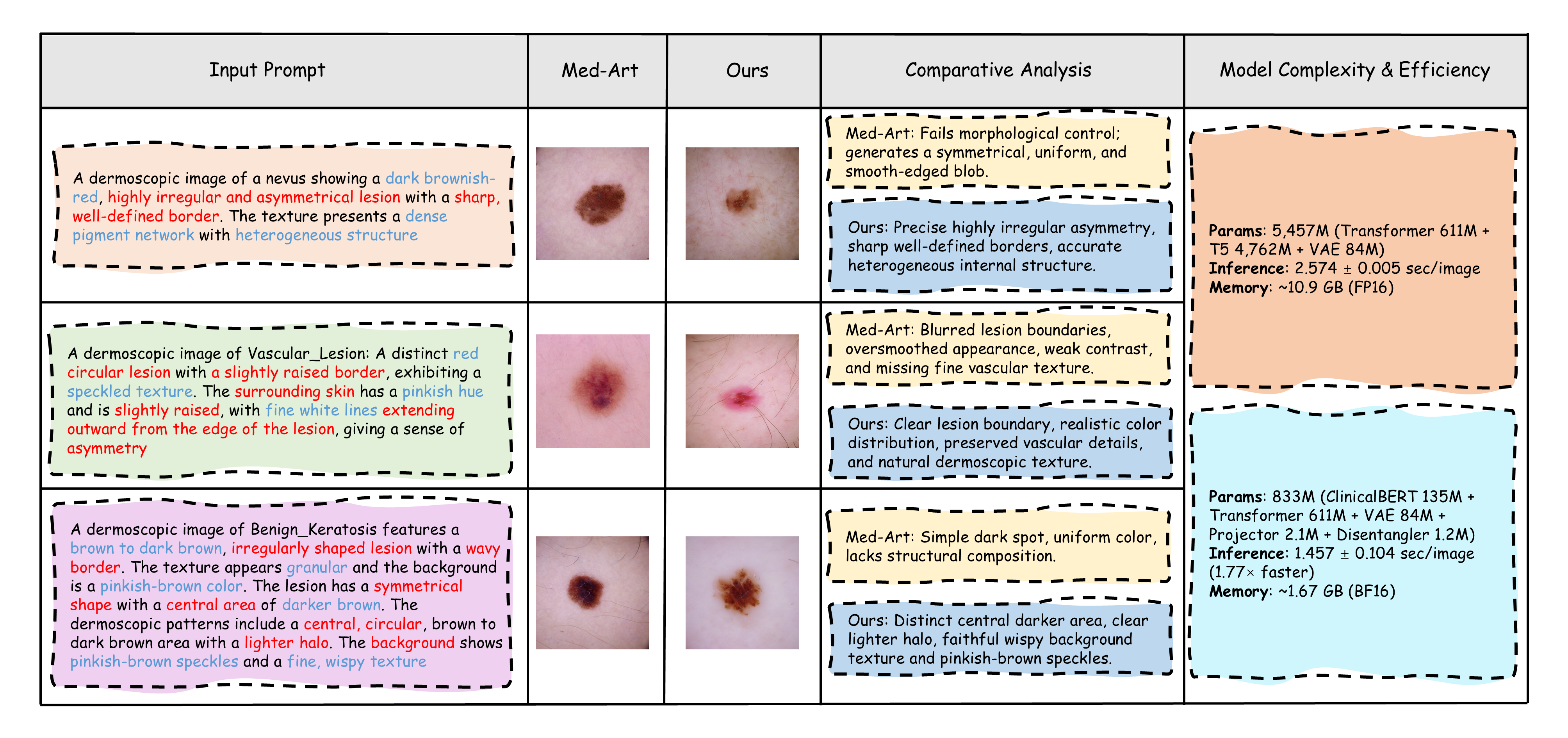}
    \caption{Impact of text prompts for comparable methods on image synthesis. Anatomical structures and modality styles are highlighted in red and blue, respectively.}
    \label{fig:motivation}
\end{figure}

Despite recent progress in adapting general-purpose text-to-image models to the medical domain \cite{chen2024medical,jiang2023anatomical}, existing approaches still exhibit limited anatomical controllability. This limitation primarily stems from the intrinsic semantic granularity gap between medical images and clinical text: images encode rich spatial and geometric details, while text is highly compressed, 
providing insufficient guidance for fine-grained structure generation. Moreover, commonly used text encoders produce global semantic embeddings in which structural and textural styles are implicitly averaged, blurring their boundaries in the latent space. 
As illustrated in Fig.~\ref{fig:motivation}, despite explicit prompts regardless of prompt lengths, details specific structural (e.g., irregular shape) and textural styles, models such as Med-Art fail to generate images with these characteristics. 
The generated images disregard both the desired irregularity and internal texture, indicating that structural cues are progressively diluted during diffusion. Such anatomically implausible generations substantially limit the effectiveness of synthetic data for improving downstream diagnostic robustness. 
Moreover, these models are computationally expensive, whereas our method reduces inference parameters to 833M (84.7\% fewer compared with Med-Art) and achieves 1.457s per image (1.77$\times$ faster), enabling efficient clinical deployment.

\begin{figure}[t]
\centering
\includegraphics[width=\textwidth]{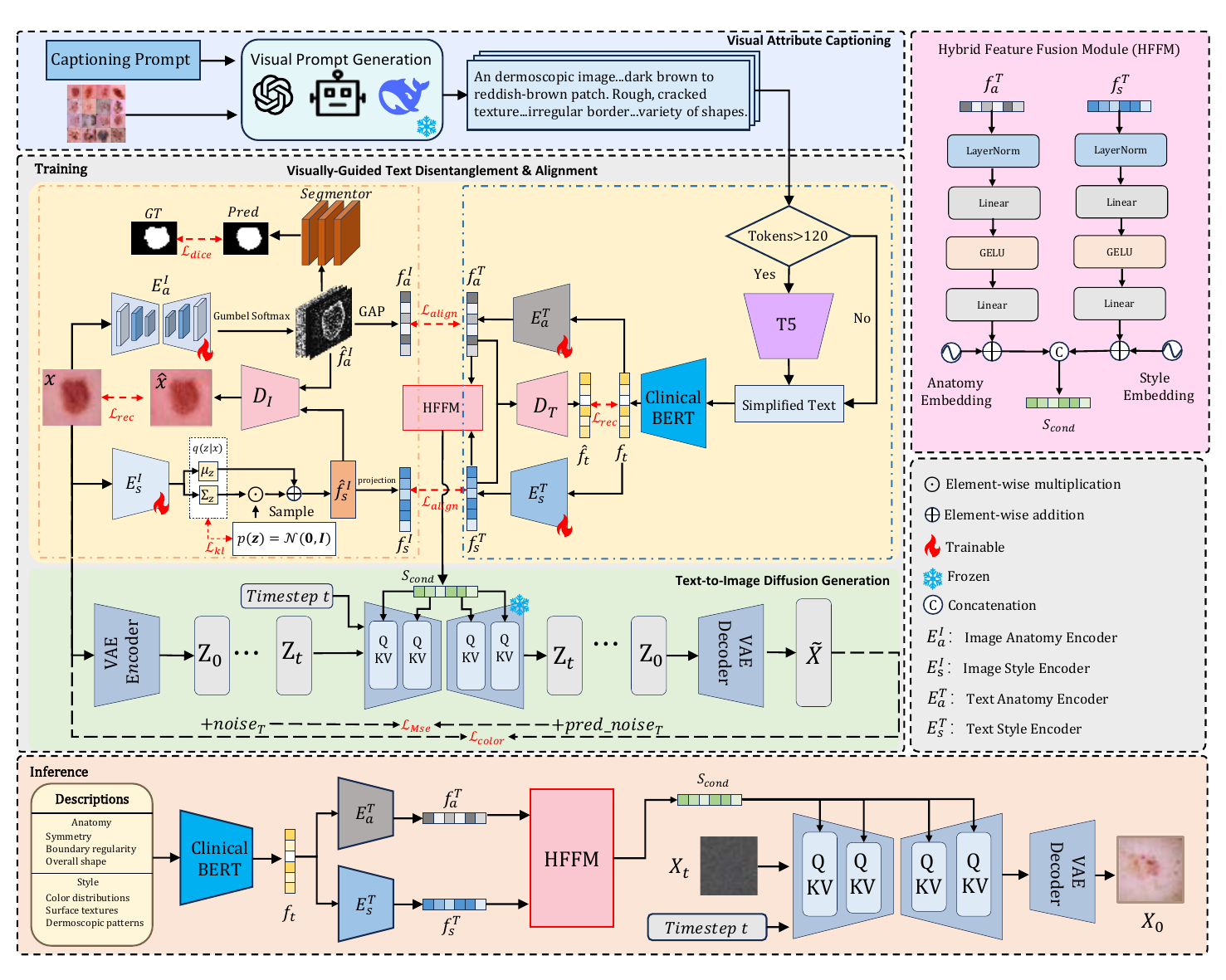} 
\caption{The overall architecture of our proposed Visually-Guided Text Disentanglement Diffusion Framework. Note that only the lightweight feature projector and the Transformer with LoRA adapters are required in the inference stage.}
\label{fig:framework}
\end{figure}

To address these issues, we propose a Visually-Guided Text Disentanglement Diffusion Framework. Our core insight is that, although textual representations are inherently abstract and susceptible to entanglement, the visual features are naturally more separable.
Accordingly, we adopt a visually guided strategy that leverages visual features as priors to constrain the learning of textual representations. Specifically, we design a cross-modal disentanglement and alignment module that introduces a pretrained visual encoder as supervision, guiding the text encoder in the latent space to explicitly disentangle input prompts into independent anatomical and style-related features. Furthermore, to ensure that these disentangled features can effectively control the generation process, we introduce a Hybrid Feature Fusion Module (HFFM), which integrates anatomical and style information through separate channels and injects them into the Diffusion Transformer (DiT). This design significantly enhances the diversity and clinical value of the generated samples without sacrificing fidelity. The main contributions of this work are summarized as follows:

\begin{itemize}
    \item We investigate whether visual guidance can enable precise semantic disentanglement, and propose to leverage visual supervision to address the semantic entanglement problem in medical T2I generation. 
    \item We design a cross-modal latent alignment strategy and a hybrid feature fusion module, enabling the generative model to learn fine-grained, biologically plausible features directly from unstructured text. Our model is designed to be lightweight, requiring fewer parameters during inference.
    \item Extensive experiments on three medical datasets demonstrate that our method outperforms existing approaches in generation quality and diversity. Notably, the synthesized data yields significant performance gains in downstream classification tasks.
\end{itemize}

\section{Method}


\subsection{Visual Attribute Captioning}
To mitigate the semantic sparsity of existing medical datasets, which lack fine-grained visual descriptions, we construct an automated attribute captioning pipeline using LLaVA-Next \cite{liu2024visual}. We design constrained prompt templates to suppress hallucinations, explicitly directing the VLM to focus on two orthogonal dimensions: Anatomy (e.g., symmetry, boundary, shape) and Style (e.g., color distribution, texture patterns). The prompt follows a structured format: "Describe the [Class] image using features: [Attribute List]". Subsequently, a T5 model \cite{raffel2020exploring} refines the generated descriptions. This process yields high-quality image-text pairs rich in structural and textural style details, providing robust priors for the subsequent cross-modal alignment.

\subsection{Visually-Guided Text Disentanglement \& Alignment}
\label{sec:method_alignment}


To resolve the high semantic entanglement between anatomical structures and modality styles in medical text, we design a visually-guided cross-modal disentanglement module. This module leverages the inherent fine-grained visual features as supervisory to guide the text encoder to map unstructured text into disentangled semantic representations in the latent space.

\noindent\textbf{Visual Disentanglement.}
We construct a dual-branch visual encoder to extract anatomical and style representations from medical images. As illustrated in Fig.\ref{fig:framework}, given an input image $x$, the Image Anatomy Encoder$E_a^I$ \cite{ronneberger2015u} adopts a U-Net-based architecture to capture spatial and geometric structures such as lesion shape and boundaries. Its output feature map $\hat{f_a^I}$ is constrained by a lightweight segmentation head with a Dice loss $\mathcal{L}_{dice}$.
This explicit supervision encourages $E_a^I$ to focus exclusively on anatomical structure. Meanwhile, the Image Style Encoder $E_s^I$ models imaging appearance attributes including texture, color, and intensity. Following a variational formulation, $E_s^I$ is implemented as a convolutional encoder that takes the concatenation of the input image $x$ and the extracted anatomical feature $\hat{f_a^I}$ as input, and maps it into a Gaussian latent distribution parameterized by $(\mu, \sigma)$. A latent style code $\hat{f_s^I}$ is then sampled via the reparameterization trick. The Kullback-–Leibler divergence loss $\mathcal{L}_{kl}$ is applied to regularize the style latent space toward a standard normal distribution, ensuring smoothness and continuity.
To maintain information integrity,
an image decoder $D^I$ is introduced to reconstruct the original image from the disentangled representations. 
The overall objective function for the visual disentanglement is defined as:
\begin{equation}
\mathcal{L}_{img} = \mathcal{L}^I_{rec} + \lambda_{dice}\mathcal{L}_{dice}+ \lambda_{kl}\mathcal{L}_{kl}
\end{equation}

\noindent\textbf{Text Disentanglement \& Cross-Modal Alignment.}
After training the visual branch, all visual encoders are frozen to serve as stable and reliable supervision signals. Given a textual description generated by a vision-language model (VLM), we first extract a base semantic embedding using a frozen ClinicalBERT encoder \cite{alsentzer2019publicly}. To endow the text representation with disentanglement capability, we design two text mapping networks: a text anatomy encoder $E_a^T$ and a text style encoder $E_s^T$. Unlike convolutional visual encoders, both $E_a^T$ and $E_s^T$ are implemented as lightweight multi-layer perceptrons, each consisting of two fully connected layers with ReLU activation and dropout. These networks operate as projection heads that map the global, entangled semantic embedding into low-dimensional latent spaces corresponding to anatomical structure and imaging style, respectively. The core of this module lies in the cross-modal latent alignment mechanism. Specifically, we enforce the text anatomical representation $f_a^T$ to align with the visual anatomical feature $f_a^I$, while the text style representation $f_s^T$ is aligned with the visual style feature $f_s^I$. This alignment is achieved by minimizing the cosine embedding distance between corresponding modality pairs. Through such explicit supervision, the text encoders learn to decompose unstructured clinical language into independent structural and appearance-related control signals. To prevent semantic loss during disentanglement, we introduce a Text Auxiliary Decoder $D_T$ to reconstruct the original text embeddings. The total objective function for this stage is defined as:
\begin{equation}
\mathcal{L}_{text} = \lambda_{a} \mathcal{D}(f_a^T, f_a^I) + \lambda_{s} \mathcal{D}(f_s^T, f_s^I) + \mathcal{L}^T_{rec}
\end{equation}

\noindent\textbf{Hybrid Feature Fusion (HFFM):} To leverage the disentangled features $f_a^T$ and $f_s^T$ for fine-grained control over the generation process, we design a Hybrid Feature Fusion Module. 
We introduce two learnable type embeddings, $e_a$ and $e_s$, as semantic identities, into the condition embedding $c_{in} = \text{concat}(\mathcal{P}_{a}(f_a^T) + e_a, \mathcal{P}_{s}(f_s^T) + e_s)$, to inject into the cross-attention layers of DiT, thereby independently guiding the synthesis of structure and style.

\subsection{Generation Process}

In the generation stage, we inject the disentangled text features into Diffusion Transformer (DiT) \cite{peebles2023scalable} and apply Low-Rank Adaptation (LoRA) \cite{hu2022lora} for fine-tuning the projection parameters of the attention layers. Furthermore, to ensure color fidelity in the generated images, we introduce an online color distribution loss $\mathcal{L}_{cd}$, constraining the pixel-level mean and variance of the generated images. The total objective function is $\mathcal{L}_{total} = \mathcal{L}_{mse} + \lambda_{cd} \mathcal{L}_{cd}$.

\begin{table}[htbp]
  \centering
  \caption{Quantitative comparison on HAM10000, Kvasir-SEG, and BUSI datasets.}
  \label{tab:comparison}
  \resizebox{\textwidth}{!}{%
  \begin{tabular}{l ccc ccc ccc}
    \toprule
    \multirow{2}{*}{\textbf{Model}} & \multicolumn{3}{c}{\textbf{HAM10000}} & \multicolumn{3}{c}{\textbf{Kvasir-SEG}} & \multicolumn{3}{c}{\textbf{BUSI}} \\
    \cmidrule(lr){2-4} \cmidrule(lr){5-7} \cmidrule(lr){8-10}
    & FID $\downarrow$ & HFD $\downarrow$ & KID $\downarrow$ & FID $\downarrow$ & KFD $\downarrow$ & KID $\downarrow$ & FID $\downarrow$ & BFD $\downarrow$ & KID $\downarrow$ \\
    \midrule
    SD1.5 \cite{rombach2022high}          & 100.05 & 15.45 & 0.078 & 119.93 & 12.34 & 0.109 & 159.74 & 62.15 & 0.125 \\
    SDXL \cite{podell2023sdxl}            & 72.69 & 8.06 & 0.058 & 139.81 & 9.78 & 0.154 & 135.20 & 55.40 & 0.098 \\
    PixArt-$\alpha$ \cite{chen2023pixart} & 68.76 & 10.14 & 0.055 & \textbf{67.73} & 5.24 & \textbf{0.048} & 100.50 & 51.94 & 0.060 \\
    MedSegFactory \cite{mao2025medsegfactory} & 52.15 & 4.85 & 0.042 & 69.80 & 4.56 & 0.058 & 99.50 & 43.10 & 0.055 \\
    Med-Art \cite{guo2025med}        & 53.54 & 9.68 & 0.049 & 70.95 & 5.80 & 0.051 & 99.16 & 43.91 & 0.062 \\
    \textbf{Ours} & \textbf{51.56} & \textbf{3.22} & \textbf{0.036} & 71.97 & \textbf{3.70} & 0.063 & \textbf{98.79} & \textbf{42.60} & \textbf{0.050} \\
    \bottomrule
  \end{tabular}%
  }
\end{table}

\section{Experiments and Results}

\subsection{Dataset and Implementation}
\textbf{Datasets.} We validated the effectiveness of our method on three public datasets. The HAM10000 dataset \cite{tschandl2018ham10000} contains 10,015 multi-source dermatoscopic images. The Kvasir-SEG dataset \cite{jha2019kvasir} includes 1,000 high-quality polyp images with precise annotations. The BUSI dataset \cite{al2020dataset} consists of 780 high-quality breast ultrasound images and corresponding masks. We randomly split the datasets into training and testing sets with an 8:2 ratio. To address the scarcity of text descriptions in the original datasets, we generated fine-grained text descriptions rich in anatomical and modal details for each image using the method described in Sec.~2.1. All images were resized to $512 \times 512$ resolution to adapt to the generative model.


\textbf{Implementation Details.}
Our framework is trained using PyTorch and the Diffusers library on a single A30 GPU. We selected PixArt-$\alpha$ ($512\times512$) as the pre-trained DiT backbone. During the fine-tuning stage, we employed the LoRA strategy (Rank=8) for parameter-efficient training, enabling BF16 mixed precision and Gradient Checkpointing to save GPU memory. The model was trained using the AdamW optimizer with an initial learning rate of $1 \times 10^{-4}$, coupled with a Cosine Scheduler with a 500-step warmup. The total training epochs are set to 15. For the color loss, we computed it every $N=500$ steps with a sampling step of $M=20$.

\begin{figure}[t]
    \centering
    \includegraphics[width=0.32\textwidth]{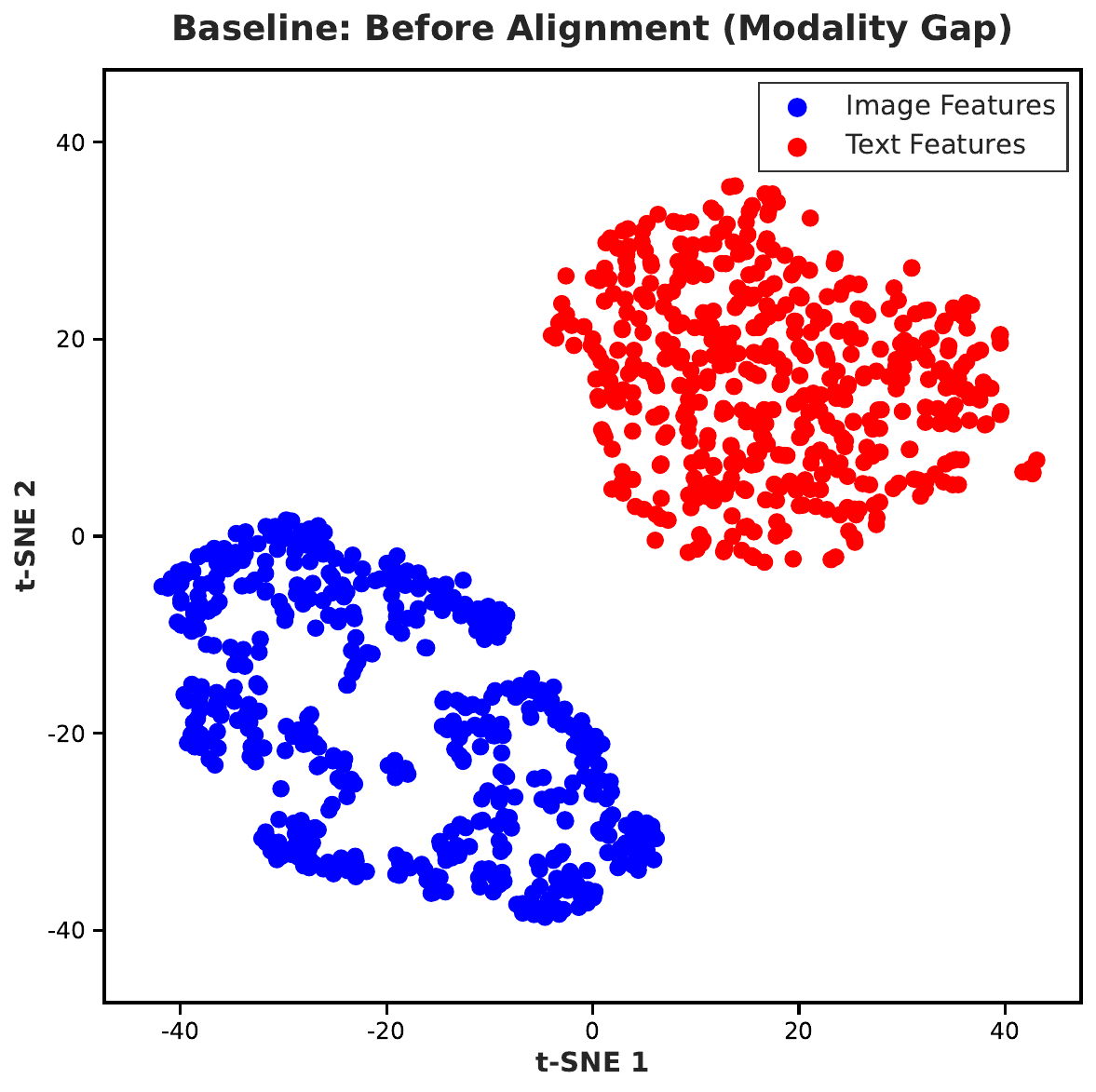}\hfill
    \includegraphics[width=0.32\textwidth]{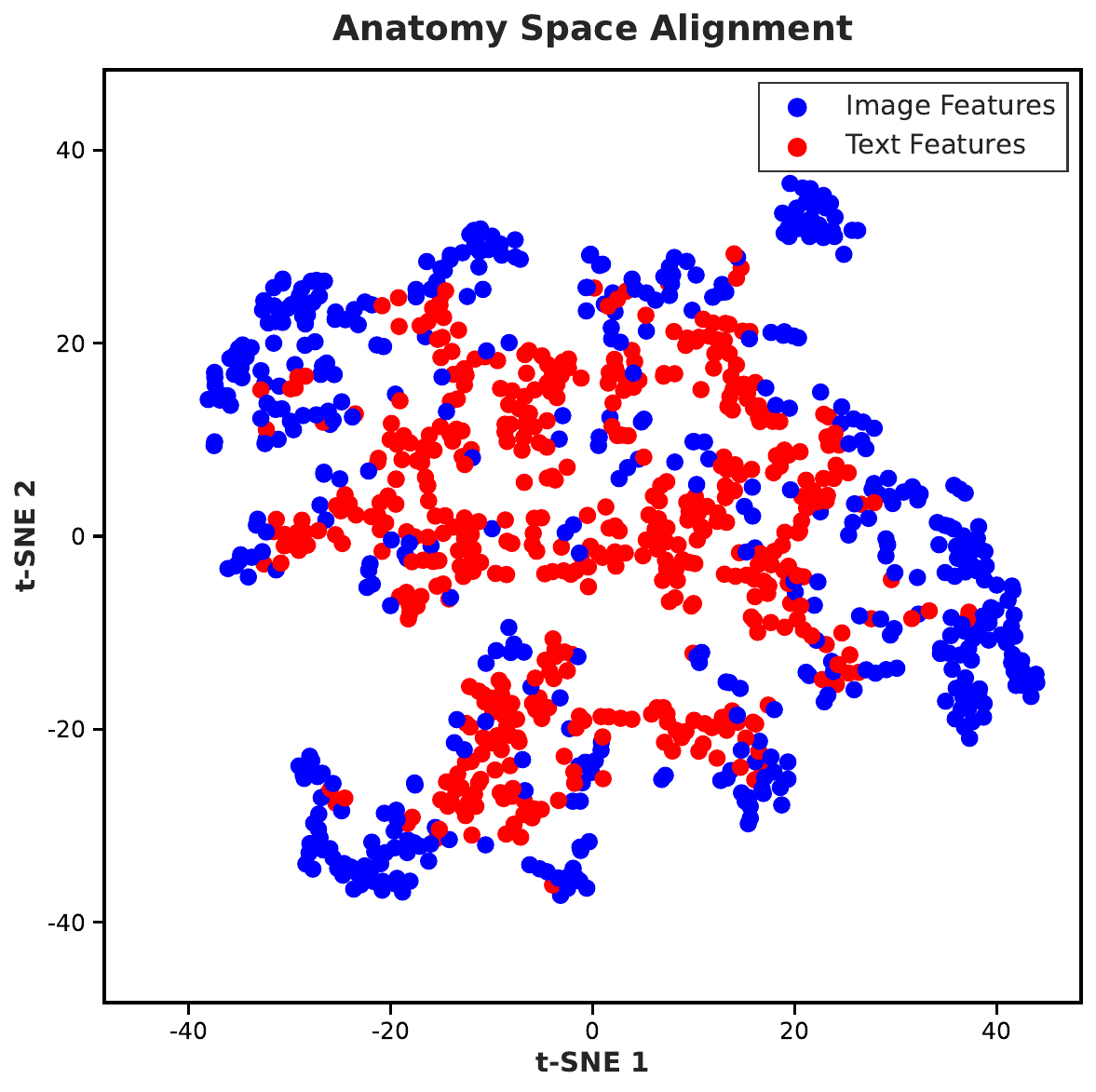}\hfill
    \includegraphics[width=0.32\textwidth]{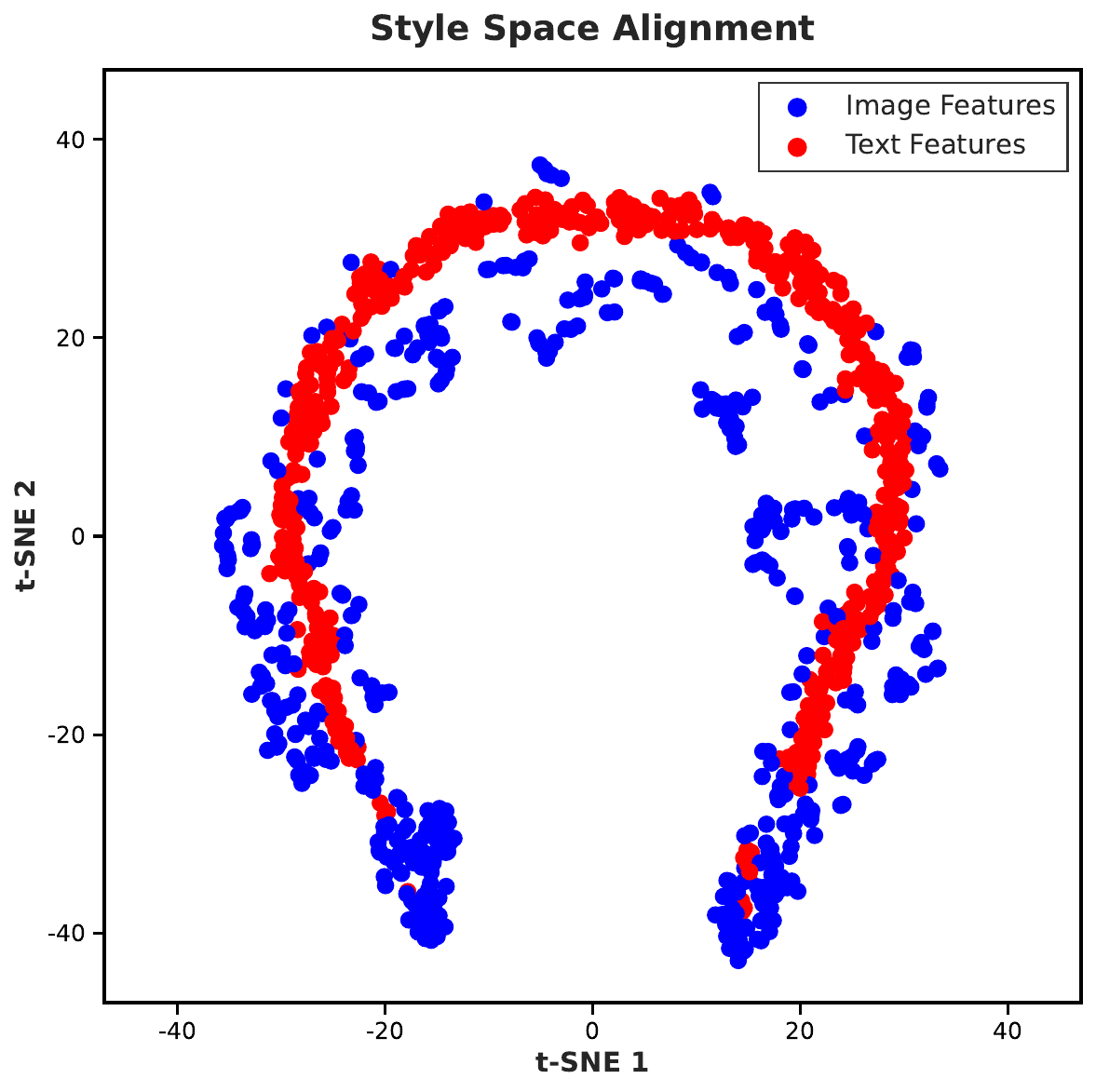}
    \caption{The t-SNE feature visualization before and after modality alignment. The baseline (Left) exhibits a significant modality gap, where text embeddings are isolated from visual features. In contrast, with our alignment module, the text features are closely aligned with the visual priors in the Anatomy (Middle) and Style (Right) subspaces.}
    \label{fig:tsne_logic}
\end{figure}

\subsection{Results}
\textbf{Evaluation of Modality Alignment.}
To intuitively verify the effectiveness of this mechanism, we present the t-SNE visualization of the feature space before and after modality alignment in Fig.~\ref{fig:tsne_logic}. It demonstrates that our model bridges the modality gap, effectively mapping unstructured text descriptions into the disentangled visual latent space.

\begin{figure*}[t] 
  \centering
  \includegraphics[width=\textwidth]{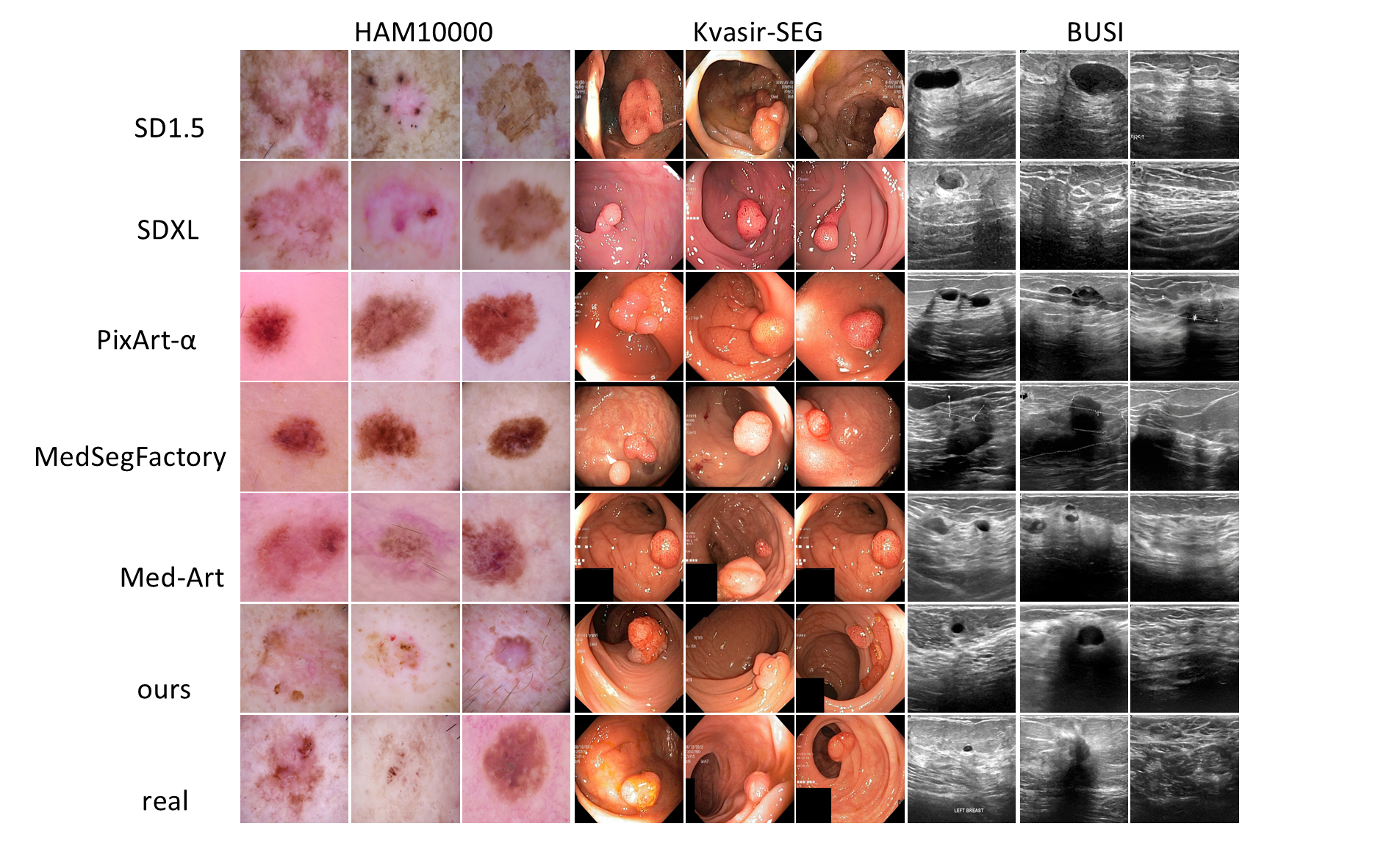} 
  \caption{Visual comparison of image synthesis results.
  }
  \label{fig:visual_comparison}
\end{figure*}

\textbf{Comparison with the Image Synthesis Methods.}
We compared our method with Med-Art \cite{guo2025med}, SD1.5 \cite{rombach2022high}, SDXL \cite{podell2023sdxl}, MedSegFactory \cite{mao2025medsegfactory} and PixArt-$\alpha$ \cite{chen2023pixart} on three datasets (Table \ref{tab:comparison}). 
Generation quality is assessed \textit{w.r.t.} FID, KID, and HFD.
On the HAM10000 dataset, our method achieves the best results across all metrics. Specifically, FID is reduced to 51.56 and HFD to 3.22, significantly outperforming the strong baseline PixArt-$\alpha$ (FID 68.76). This clearly demonstrates the compelling capability of the visually-guided disentanglement strategy in capturing fine-grained medical features. For the more challenging Kvasir-SEG and BUSI datasets, our method also exhibits competitive performance. Notably, our method consistently maintains the lowest HFD scores (e.g., 3.70 on Kvasir-SEG, better than 5.24 of PixArt-$\alpha$). This indicates that even when dealing with complex domain gaps, our approach retains a significant advantage in preserving high-frequency details crucial for clinical diagnosis (e.g., mucosal textures and lesion boundaries).
Furthermore, as shown in Fig.~\ref{fig:visual_comparison}, compared to baseline models like SDXL, images generated by our method are closer to real samples in terms of textural realism and structural consistency. Particularly in dermoscopic images, our model accurately produces complex medical features, such as clearly visible hair details and irregular pigment networks, whereas these subtle features are often over-smoothed or lost in the results of other models. 
We further compare the model sizes of different methods. Compared with Stable Diffusion 1.5 (1.07B parameters), SDXL (7.5B), PixArt-$\alpha$ (4.9B), and MedSegFactory (1.425B), our method requires only 0.833B parameters, achieving a 22\% reduction even compared to the smallest baseline.

\textbf{Downstream Classification Tasks.}
To verify the clinical utility of the synthetic data, we used the generated images to augment the HAM10000 dataset for training a classifier. Our method achieves the best F1 score (0.6185) and BACC (0.3475) in Table \ref{tab:downstream}, respectively. Although SDXL shows a slight advantage in AUC, our method's higher F1 and BACC indicate that our generated samples contain richer discriminative features. 
This confirms that our approach enhances the robustness of diagnostic models and significantly narrows the gap between synthetic and real data.

\textbf{Ablation Study.}
We evaluated the effectiveness of key components on the HAM10000 dataset (Table \ref{tab:ablation}). To validate the effectiveness of text guidance, we compared text descriptions (attribute captioning) with only class labels as guidance. After removing attribute captioning, FID significantly deteriorated from 51.56 to 69.48. Notably, naive feature concatenation (Naive Feature Concat.) performs even worse than using only class labels (FID 86.24), suggesting that disordered high-dimensional features introduce noise. In contrast, the combination of HFFM and visually-guided disentanglement achieves an FID of 51.56, validating the importance of the structured alignment strategy in generating high-fidelity medical images.

\begin{table}[htbp]
  \centering
  \begin{minipage}[t]{0.48\textwidth}
    \centering
    \caption{Downstream classification performance on HAM10000.}
    \label{tab:downstream}
    \resizebox{\linewidth}{!}{%
      \setlength{\tabcolsep}{2.5pt}
      \begin{tabular}{l ccc}
        \toprule
        \textbf{Model} & \textbf{F1} $\uparrow$ & \textbf{BACC} $\uparrow$ & \textbf{AUC} $\uparrow$ \\
        \midrule
        SD1.5           & 0.595 & 0.304 & 0.813 \\
        SDXL            & 0.558 & 0.339 & \textbf{0.841} \\
        PixArt-$\alpha$ & 0.536 & 0.143 & 0.443 \\
        MedSegFactory   & 0.598 & 0.345 & 0.816 \\
        Med-Art         & 0.515 & 0.333 & 0.806 \\
        \textbf{Ours}   & \textbf{0.619} & \textbf{0.348} & 0.830 \\
        \midrule
        Real Data & 0.816 & 0.657 & 0.964 \\
        \bottomrule
      \end{tabular}%
    }
  \end{minipage}
  \hfill
  \begin{minipage}[t]{0.50\textwidth}
    \centering
    \caption{Ablation study of different components on HAM10000.}
    \label{tab:ablation}
    \resizebox{\linewidth}{!}{%
      \setlength{\tabcolsep}{2pt} %
      \begin{tabular}{l ccc}
        \toprule
        \textbf{Configuration} & \textbf{FID} $\downarrow$ & \textbf{HFD} $\downarrow$ & \textbf{KID} $\downarrow$ \\
        \midrule
        w/o Attr. Captioning & 69.48 & 11.34 & 0.059 \\
        Naive Feature Concat. & 86.24 & 8.84 & 0.073 \\
        w/o Disentanglement & 53.54 & 9.68 & 0.049 \\
        \midrule
        \textbf{Ours} & \textbf{51.56} & \textbf{3.22} & \textbf{0.036} \\
        \bottomrule
      \end{tabular}%
    }
  \end{minipage}
\end{table}

\section{Conclusion}
In this work, we propose a Visually-Guided Text Disentanglement Diffusion Framework for medical T2I generation. By leveraging cross-modal latent alignment, visual priors are disentangled into anatomical and stylistic representations, which are injected into a DiT backbone via a Hybrid Feature Fusion Module (HFFM) for fine-grained control. 
Extensive experiments show compelling generation fidelity and diversity, while the generated synthetic data significantly improves downstream diagnostic performance.

%
%
\bibliographystyle{splncs04}
\bibliography{ref}
\end{document}